\pdfoutput=1
\documentclass[11pt]{article}

\usepackage[]{emnlp2021}

\usepackage{times}
\usepackage{latexsym}

\usepackage[T1]{fontenc}
\usepackage[utf8]{inputenc}

\usepackage{microtype}

\usepackage{enumitem}
\usepackage[official]{eurosym}

\usepackage{graphicx}
\usepackage{xcolor,colortbl}
\usepackage{amsmath}
\usepackage{booktabs}
\usepackage{multirow}
\usepackage{comment}
\usepackage{bm}
\usepackage{microtype}
\usepackage{amssymb}% http://ctan.org/pkg/amssymb
\usepackage{pifont}% http://ctan.org/pkg/pifont.bib

\usepackage[font=small,labelfont=bf]{caption}
\usepackage{subcaption}

\usepackage{array}
\newcolumntype{P}[1]{>{\centering\arraybackslash}p{#1}}
\newcolumntype{M}[1]{>{\centering\arraybackslash}m{#1}}

\definecolor{nicegreen}{HTML}{009B55}

\title{What Vision-Language Models `See' when they See Scenes} %
% ...

\author{Michele Cafagna$^1$ \ \ ~~Kees van Deemter$^2$ \ \ ~~Albert Gatt$^{1,2}$\\\\
$^1$University of Malta, Institute of Linguistics and Language Technology \\
$^2$Universiteit Utrecht, Information and Computing Sciences \\ 
\texttt{michele.cafagna@um.edu.mt}\\\texttt{\{c.j.vandeemter,a.gatt\}@uu.nl}
}

\date{}

\begin{document}
\maketitle

\begin{abstract}
%Multimodal Leaning focuses on trying to use different kinds of information for improving the performance of a model. In our context, it is designed to leverage two sources of information, namely visual and textual. Vision and language models are pre-trained on an image-sentence objective, which is considered beneficial to acquire grounding capabilities. This task consists in telling if a caption is aligned or related to the image. However, the captions in current datasets are usually very descriptive or low-level, and there is a lack of generalization of abstract concepts or high-level. Moreover, the grounding capabilities are never deeply explored, indeed these models are judged on a downstream task benchmark scores basis. 
% Multimodal grounding --- the ability of models to incorporate perceptual features in meaning representations --- has received new impetus through the development of large-scale, pretrained, general purpose Vision and Language (V\&L) models. Such models are typically trained on large datasets of images paired with texts using an image-sentence matching task, among others. 
Images can be described in terms of the objects they contain, or in terms of the types of scene or place that they instantiate. In this paper we address to what extent pretrained Vision and Language models can learn to align descriptions of both types with images.
% Pretrained Vision and Language (V\&L) models are usually trained on images with captions which describe entities and their spatial relationships. In addition to such `object-level' descriptions, it is also possible to describe images in terms of the overall scene they depict. In this paper, we compare the ability of V\& L models to match scene-level and object-level captions to their respective images, 
We compare 3 state-of-the-art models,
% We investigate the ability of pretrained Vision and Language (V\&L) models to reason about the connection between captions which describe entities and their spatial relationships, versus `scene-level' captions, which capture generic information about scenes.
%In many V\&L datasets, the texts are usually highly descriptive, emphasising the entities visible in an image and their spatial relations. This raises the question whether V\&L models are capable of learning more general semantic relationships. In this paper, we consider the difference between `low-level' descriptive text and higher-level descriptions that capture generalisations about scenes.
% The question we address is whether such models are able to establish a link between entity-level descriptions and high-level scene descriptions.
% We address this question with three 
% Out of three state-of-the-art vision and language (V\&L) models, 
VisualBERT, LXMERT and CLIP. We find that (i) V\&L models are susceptible to stylistic biases acquired during pretraining; 
% KvD Where in the paper do we discuss this first finding? Sorry for overlooking it.
(ii) only CLIP performs consistently well on both object- and scene-level descriptions. A follow-up ablation study shows that CLIP uses object-level information in the visual modality to align with scene-level textual descriptions.
\end{abstract}

\section{Introduction}\label{sec:intro}
Grounding symbols in perception \cite{Harnad1990} is a crucial step towards achieving full understanding of natural language \cite{bender-koller-2020-climbing,bisk-etal-2020-experience}. This endeavour has received new impetus through the development of pretrained Vision and Language (V\&L) models  \cite[e.g.][]{lu2019vilbert,tan-bansal-2019-lxmert,li2019visualbert,chen2020uniter,Li-etal-2020unicodervl,Su2020VL-BERT,Wang2020,Luo2020,Li2021,Huang2021,Radford2021}. Similarly to unimodal language models such as BERT \cite{devlin2019bert}, V\&L models are intended to be task-agnostic and are extensively pretrained on paired image-text data, achieving good performance on several tasks after finetuning  \cite[e.g.][]{lu2020vilbert12in1, li2020oscar, kim2021vilt}. Pretraining usually includes an image-text alignment task to discover implicit cross-modal relationships. Although the importance of this task is widely recognized and adopted during model pretraining, 
% KvD say: the pre-training of what?
it is unclear how the models perform on it, since they are usually evaluated on downstream tasks.
% A common pretraining task is image-text alignment, whereby a model needs to determine whether an image and a text correspond; this is usually coupled with other pretraining objectives, including masked language modelling and masked object/region prediction. 
% These models have been shown to 

% \begin{minipage}[b]{0.65\linewidth}
% some text
% \end{minipage}
% \hfill
% \begin{minipage}[b]{0.35\linewidth}
% \includegraphics[height=8\baselineskip]{grafics}
% \end{minipage}

\begin{figure}[!t]
    \begin{minipage}[b]{\linewidth}
        \centering
        \includegraphics[scale=0.35]{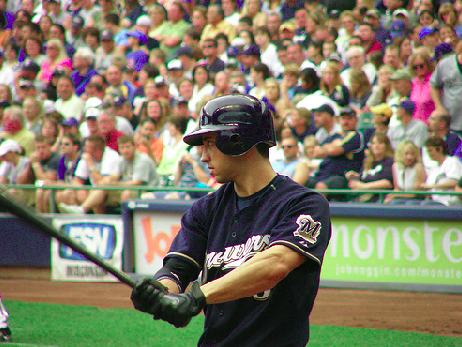}
    \end{minipage}
    % \vspace{5mm}
    \begin{minipage}[b]{\linewidth}
        \footnotesize
        {\bf LN:} This is the picture of a stadium. In the foreground there is a person [\textellipsis] At the back there are group of people sitting [\textellipsis].\\
        {\bf COCO:} a baseball player getting ready to swing at a baseball game in a stadium packed with people. \\
        {\bf HL1K:} the picture is shot in a baseball field
    \end{minipage}
    \caption{An example of scene with COCO and Localized Narrative (LN) object-level captions, versus HL1K scene-level description (Section \ref{sec:data})}
    \label{fig:example-image}
\end{figure}

% Good performance and transferrability raise a number of questions concerning the precise nature of these models' representational and reasoning abilities. 
% In this paper, we address the extent to which V\&L models learn to link relatively low-level descriptive information about visual scenes, to higher-level information. 
The data used 
% Many of the datasets used 
for V\&L pretraining usually contains highly descriptive text which mentions objects and their spatial relationships. For instance, the COCO \cite{Chen2015} and Localized Narratives \cite[LN;][]{Pont-Tuset2019} captions for Figure \ref{fig:example-image} are of this type, though they differ stylistically. By contrast, the third caption in the figure, from the novel HL1K dataset introduced in Section \ref{sec:data} below, is what we refer to as `scene-level', focusing on {\em what type} of scene or location is depicted. 
%%AG: BELOW IS NEW, provides a bridge to the following par, and motivates some of the work in this paper
Note that both the object- and scene-level descriptions in the Figure describe the picture, albeit in different ways. Indeed, it would be expected that, for a V\&L model to display true grounding capabilities, it should be able to identify both types of descriptions as true of a scene. For models which do display this cabability, a natural follow-up question is whether their representations capture interesting connections between scenes on the one hand, and the objects within them on the other.

%%AG: significant changes to the below par, wiht more research cited
Research on human perception suggests that humans do not perceive scenes exclusively in terms of the objects they contain, and that visual salience is not exclusively determined by bottom-up features such as colour and texture. Rather, visual stimuli are considered `scenes' because their elements constitute a meaningful whole, both in terms of their contents (e.g. one expects an oven in a kitchen, but not in a living room) and in terms of their spatial arrangement (e.g. ovens do not typically hang from the ceiling) \cite{Malcolm2016}. These observations gave impetus to work showing that violations of scene `semantics' (content) and `syntax' (spatial arrangement) exact a cognitive cost during perception \cite[e.g.][]{Biederman1982,Vo2013}. A related strand of modeling research in computer vision has also shown that scene-level priors generate expectations about objects and their configurations, impacting the salience of objects in a way that classical, feature-based models of attention \cite[e.g.][]{Itti2001} would not predict \cite{Torralba2006,Oliva2007}. Indeed, the problem of linking low-level features with high-level semantic information is an instance of the problem referred to as the `semantic gap' in computer vision \cite{Ma2010}.

%%%% NEW PART
In this paper we investigate whether V\&L models are able to handle object-level and scene-level descriptions equally well. A positive answer to this question would suggest that such models are learning useful associations between the elements of a scene and the overall scene type, as captured in the textual descriptions.

We perform an analysis in a zero-shot setting on three state-of-the-art pretrained Vision and Language models. To our knowledge, this is the first systematic comparison of model capabilities on object- versus scene-level grounding.
%Since we have no previous experience regarding the model’s capability to ground such relationships, in our explorative study, we choose 
%%%AG: Minor edits below
The goal of this study is therefore not to establish new SOTA results, but to further our understanding of what V\&L models learn, as a function of the data they are pretrained on and the model architecture. 
%we do not focus on finding the best performing model, in fact we aim at extending the exploration to a wider range of already available pretrained V\&L models. 
Therefore we choose three models differing in many settings, (including training set size, architecture, number of parameters and model size). All of the models are however optimized on the image-sentence alignment task.

We find that only one of the models under comparison, CLIP \cite{Radford2021}, performs consistently well on both object- and scene-level image-text matching. We then investigate this model's abilities in depth, using an ablation method to identify the elements of a text and/or an image which contribute to these abilities.  

% KvD If we need to save space, I think the entire next paragraph can go.
% The main contributions of this paper are (a) the development of an ablation method on both visual and textual information to probe grounding capabilities; (b) a comparison of V\&L models on image-sentence matching with object- and scene-level descriptions; (c) an analysis of model capability to link descriptive elements to higher-level concepts; (d) the introduction of a new dataset designed for research into object-level versus scene-level image-text understanding.

\section{Models}\label{sec:models}
Current V\&L models typically combine textual and visual features in a single or a dual-stream architecture. 
Though the two architectures have been found to perform roughly at par \citet{bugliarello2020multimodal}, in this paper we include representatives of both. We also include a third model which differs in structure and is trained on a much larger and more varied dataset. Table \ref{table:models-comparison} gives an overview of some of the properties of the models we consider.

\begin{table}[t]
    \small
    \centering
    \resizebox{\columnwidth}{!}{
    \begin{tabular}{c||c|c|c}
        \toprule
        & \multicolumn{1}{c|} {Training size}
        & \multicolumn{1}{c|} {Model size} 
        & \multicolumn{1}{c} {Pretraining } \\
        & \multicolumn{1}{c|} {(\# image-sentence pairs)}
        & \multicolumn{1}{c|} {(\# parameters)} 
        & \multicolumn{1}{c} {Objectives}\\
        \midrule
        CLIP & 400M & 151M & ISA \\
        \midrule
        VisualBERT & 330k & 112M & ISA, MLM \\
        \midrule
        \multirow{2}{*} {LXMert} & \multirow{2}{*} {9.18M} & \multirow{2}{*} {228M} & ISA, MLM \\
        & & & MOP, VQA \\
        \bottomrule
    \end{tabular}
    }
    \caption{Comparison of training settings for the three models (\textbf{ISA}: Image-Sentence Alignment, \textbf{MLM}: Masked Language Modeling, \textbf{MOP}: Masked Object Prediction, \textbf{VQA}: Visual Question Answering) }
     \label{table:models-comparison}
\end{table}

\paragraph{LXMERT} \cite{tan-bansal-2019-lxmert} is a dual-stream model, which encodes text and visual features in parallel, combining them using cross-modal layers. LXMERT is trained on COCO captions \cite{Chen2015} as well as a variety of VQA datasets, with an image-text alignment objective, among others. 

\paragraph{VisualBERT} \cite{li2019visualbert} is a single-stream, multimodal version of BERT \cite{devlin2019bert}, with a Transformer stack to encode image regions and linguistic features and align them via self-attention. It is pretrained on COCO captions \cite{Chen2015}. Image-text alignment is conceived as an extension of the next-sentence prediction task in unimodal BERT. Thus, VisualBERT expects an image and a correct caption, together with a second caption, with the goal of determining whether the second caption matches the $\langle$image, correct caption$\rangle$ pair.

\paragraph{CLIP} \cite{Radford2021} combines a transformer encoder for text with an image encoder based on Visual Transformer \cite{Dosovitskiy2020}, jointly trained using contrastive learning to maximise scores for aligned image-text pairs. CLIP is trained on around 400m pairs sourced from the Internet, a strategy similar to the web-scale training approach used for unimodal models such as GPT-3 \cite{Brown2020}. We note that the visual backbone for this model differs from that of LXMERT and VisualBERT, both of which use Faster-RCNN \cite{Ren2015}.

\section{Data}\label{sec:data}
We use four different datasets for our experiments. Dataset statistics are provided in the Appendix.

\paragraph{Localized Narratives} Localized Narratives (LN) \citet{Pont-Tuset2019} is a V\&L dataset harvested by transcribing speech from annotators who were instructed to give object-by-object descriptions as they moved a mouse over image regions. LN captions tend to be highly detailed and stylistically similar to speech. We use LN as a source of object-level captions. 
The images in LN come from preexisting datasets; this allows us to align LN captions with images and captions from datasets such as COCO and ADE20K.

\paragraph{ADE20K} ADE20K \cite{zhou2017scene} is a computer vision dataset containing 20k images comprehensively annotated with objects, parts and scene labels. We use ADE20K as a source of scene-level captions. For our experiments, we filter out images with scenes which in the dataset are labelled as \texttt{unknown}. We produce captions for each image using a simple template-based generation method (see Appendix).
We align ADE20K images and their scene-level descriptions, to the corresponding object-level captions in LN.

\paragraph{COCO} COCO \cite{lin2014microsoft} consists of images paired with captions and object annotations. 
LN captions are also available for the same images.
We use images and captions from the 2017 COCO validation split, as well as the corresponding LN captions.

\paragraph{HL1K} {\bf H}igh {\bf L}evel Scenes - 1k (HL1K) is a new dataset and part of an ongoing data collection effort. 
HL1K is composed of 1k images, each depicting at least one person, sampled from the 2014 COCO {\tt train} split. We crowd-sourced three annotations per image on Amazon Mechanical Turk, showing crowd workers the image and asking them to write a description in response to the question \textit{Where is the picture taken?} It was made clear to annotators that the answer to this question should bring to bear their knowledge of typical, or common, scenes.
Descriptions were corrected for typos using the Neuspell Toolkit \cite{jayanthi2020neuspell}.
Finally, we paired our scene-level HL1K captions with the previously available COCO and LN object-level captions. Figure \ref{fig:example-image} is an example; further examples are provided in the Appendix.

\section{Experiments}
We first test models in the image-text alignment task on both object- and scene-level descriptions. We then delve more deeply into the performance of the best model, using an ablation task. 
Since we are interested in the capabilities of the pretrained models, we do not finetune them.

\subsection{Image-sentence alignment}

\begin{table}[t]
    \small
    \centering
    \begin{tabular}{ll||ccc}
    \toprule
                                   &             & {\sc lxmert} & {\sc clip} & {\sc v}isual{\sc bert}\\
    \midrule
     \multirow{3}{*}{Object} & ADE20k + LN & 28.4 & \bf 96.8 & 39.0 \\
                                   & COCO + LN   & 59.1 & \bf 98.7 & 65.2 \\
                                   & COCO Cap.   & 79.3 & \bf 99.1 & 64.4 \\
    \midrule
    \multirow{2}{*}{Scene}   & ADE20k      & 58.0 & \bf 97.6 & 17.3 \\
                                   & HL1K        & 45.5 & \bf 91.5 & 55.3 \\
    \bottomrule 
    \end{tabular}
    \caption{Image-sentence alignment accuracies on object-level and scene-level captions. Chance performance is at 50\%. ({\em LN} = Localized Narratives)}
    \label{table:alignment}
\end{table}

For the image-text alignment task, we use the models' pretrained alignment head to predict whether a scene-level or object-level caption correctly describes an image, or not.\footnote{Note that this setting is the same used by the models is their  pretraining.} Table \ref{table:alignment} shows that LXMERT and VisualBERT perform adequately on object-level COCO Captions, though performance is lower than expected given that they were pretrained on this dataset. In the case of LXMERT, one possible explanation is catastrophic forgetting, arising from the fact that this model is pretrained for its final ten epochs on VQA \cite[similar observations are made by][]{Parcabalescu2021}. For both models, performance drops dramatically on LN captions. This is likely due to a stylistic difference: LN captions are longer than COCO, more discursive and contain disfluencies. In contrast, CLIP performs close to ceiling on all three datasets, possibly 
reflecting the benefits accrued from the size and diversity of its pretraining data.

On scene-level captions, performance is somewhat above chance for LXMERT on ADE20k template-based descriptions, and for VisualBERT on HL1K. Otherwise, performance is below 50\% for both models. Once again, CLIP performs above 90\%, though there is a drop in performance from the template-based ADE20k descriptions to human-authored HL1K scene-level captions, possibly due to the more predictable nature of the former.

\subsection{Ablation experiments on CLIP}
%%AG: Gave some justificaiton for this part immediately below.
Since CLIP is the only one of the three models which is successful at matching scene-level and object-level captions to images, we probe its capabilities further, paying particular attention to the question whether CLIP links {\em scene types} (e.g. kitchen) to {\em scene contents} (e.g. oven, pizza) in image-text matching.
% In what follows, we probe the capabilities of CLIP further.

Whereas a standard image-text alignment setup  
compares the model's success at identifying actual versus random captions, here we directly compare the preference of the model for scene- versus object-level descriptions, as a function of (i) the entities mentioned in the object-level caption; (ii) the entities visible in an image. To this end, we use textual and visual ablation, described below (see Appendix for further examples and details).

\paragraph{Textual ablation} A textual ablation is performed by removing noun phrases (NPs) in the object-level caption. For example, for the COCO caption in Figure \ref{fig:example-image}, one ablated version would remove the NP {\em a baseball player} resulting in {\em getting ready to swing at a baseball game \textellipsis}. For a given image $i$ with object-level caption $o$ and scene-level caption $s$, we compare how $P(o \vert i)$ -- CLIP's estimate of the probability that $o$ matches $i$ -- changes as NPs are removed from $o$, and to what extent this causes CLIP to assign higher probability $P(s \vert i)$, to $s$ as the match for $i$. We report two comparisons, one on LN captions versus ADE20K scene template descriptions; and one on COCO captions against HL1K scene-level descriptions. 

To control for possible loss of grammaticality after ablation, we score ablated captions with GRUEN \cite{Zhu2020}, a BERT-based model which has been shown to yield scores that correlate highly with human judgments.\footnote{GRUEN returns a combined score consisting of a linear combinaton of Grammaticality, Focus and Coherence. Here, we use only the Grammaticality scores.} CLIP probabilities for ablated textual captions yielded a very low correlation with grammaticality (Pearson's $r=0.1, p < .01$) suggesting that grammaticality did not affect the scores.

\paragraph{Visual ablation} Given an image, we detect the bounding regions of entities which are also mentioned in the corresponding object-level caption, and occlude them with a greyscale mask. Once again, we are interested in whether CLIP's estimate of the alignment probability of object- versus scene-level captions, changes as elements of the visual input are masked.

\begin{table}[t!]
    \small
    \centering
    \begin{tabular}{c||cc|cc}
        \toprule
        & \multicolumn{2}{c|} {ADE20k} & \multicolumn{2}{c} {HL1K} \\
        \midrule
        & LN & Scene & COCO & Scene \\
        \cmidrule{2-5}
        No ablation & 55.6 & 44.4 & 95.7 & 4.3 \\
        \cmidrule{2-5}
        T & 22.0 & 78.0 & 67.2 & 32.8 \\
        \cmidrule{2-5}
        V & 74.9 & 25.1 & 71.2 & 28.8 \\
        \cmidrule{2-5}
        V+T & 68.4 & 31.6 & 63.3 & 36.7 \\
        \bottomrule
        % \midrule
        % & COCO cap.. & HL cap. \\
        % No ablation  & 95.7 & 4.3 \\
        % \cmidrule{2-3}
        % T & 67.2 & 32.8 \\
        % \cmidrule{2-3}
        % V & 71.2 & 28.8 \\
        % \cmidrule{2-3}
        % V+T & 63.3 & 36.4 \\
        % \cmidrule{2-3}
    \end{tabular}
    \caption{Preference expressed in percentage for object-level versus scene-level (sub-columns), where there is no ablation, T(extual) ablation, V(isual) ablation, or both (V+T) for a each dataset (columns).}
     \label{table:ablation-results}
\end{table}

\paragraph{Vision vs text} Without ablation, 
the model assigns higher probability to object-level descriptions, suggesting that CLIP assigns higher confidence to an image-text pair when the text focuses on objects. This preference is far more marked for COCO/HL1K, in line with the observation (Table \ref{table:alignment}) that HL1K scene descriptions are somewhat more challenging for this model. As entity-level information is removed from the object-level caption (row T), the model's preference swings to the scene-level caption, suggesting that the model leverages the visual information to align with the scene description. Visual ablation (V), by contrast, results in the opposite tendency: when entities are occluded in the image, the model shows greater preference for object-level captions. 
% KvD should this be said more precisely? Something like "as further objects are occluded, the model shows greater preference for object-level captions"?
This suggests that the way CLIP `understands' scene-level descriptions depends on which entities are visible. If both sources of information are ablated (V+T), the preference once again veers towards object-level captions.

% KvD This is quite interesting, isn't it?

\begin{figure}[t]
    \begin{minipage}[t]{0.45\linewidth}
        \small
        \begin{tabular}{c|c}
        \toprule
        {\sc entity} & $P(s \vert e)$ \\
        \midrule
             skateboard & 0.28 \\
             bench & 0.22 \\
            frisbee & 0.11 \\
            kite & 0.11 \\
            orange & 0.11 \\
            umbrella & 0.09 \\
            person & 0.07 \\
        \bottomrule
        \end{tabular}
        \captionof{table}{Top entities in {\em park} scenes.}
        \label{table:cond-prob}
    \end{minipage}
    \hfill
    \begin{minipage}[t]{0.5\linewidth}
        \centering
        \centerline{\includegraphics[scale=0.35]{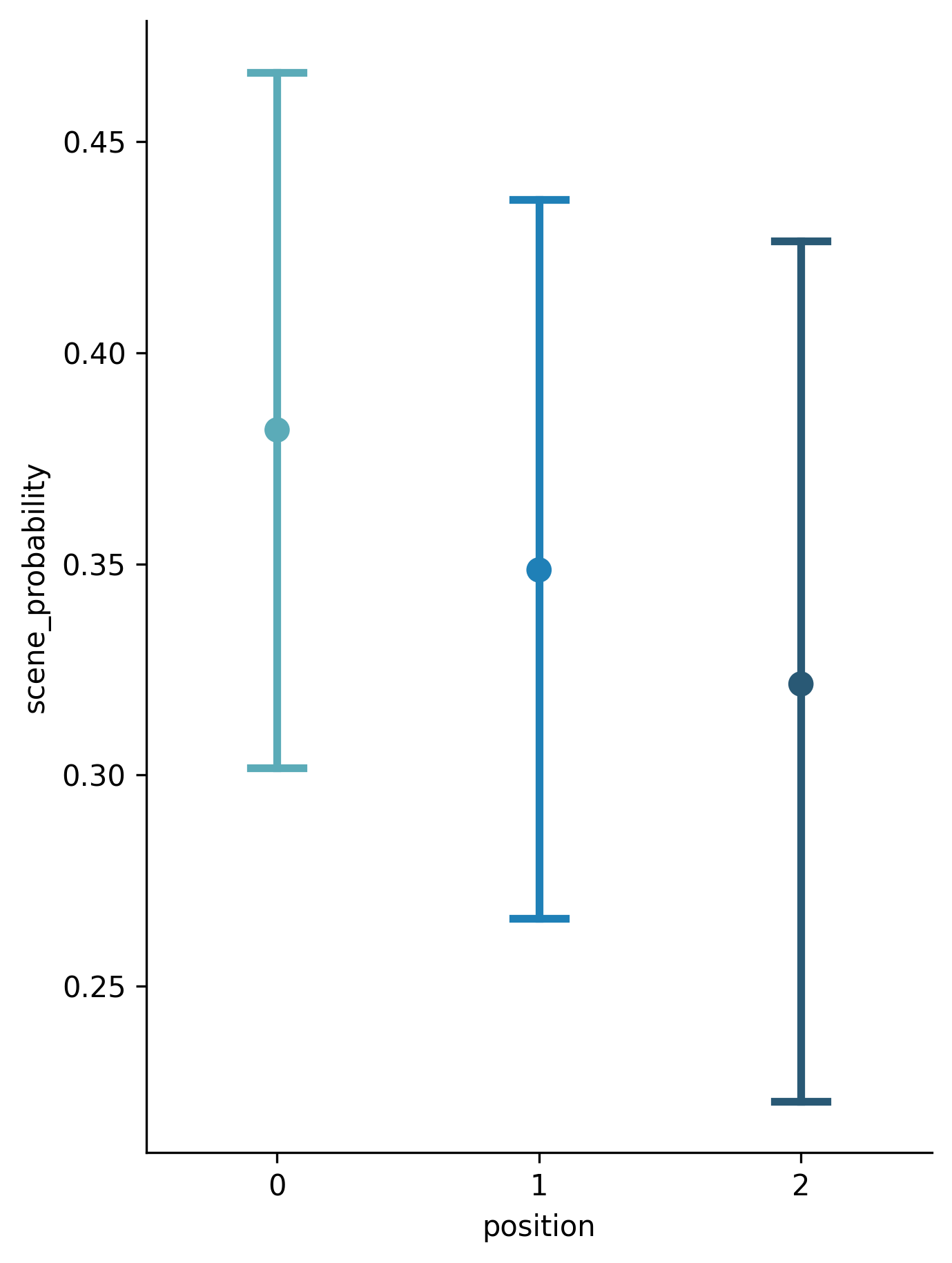}}
        \captionof{figure}{CLIP scene-level description probabilities after masking top 1-3 entities. Error bars represent standard deviations.}\label{fig:scene-prob}
    \end{minipage}
% \centerline{\includegraphics[scale=0.15]{imgs/top_3_scene_probs.png}}
\end{figure}

\paragraph{Scenes vs. entities} 
%%AG: I have reinstated the initial sentence here. We can compress for the submission, but for Arxiv, it's fine
These findings suggest that CLIP reasons about scenes on the basis of salient objects within them. To investigate this further, we use scene labels extracted from the HL1K captions and  the object detections produced for the visual ablation. For a scene label $s$ and entity label $e$, we compute $P(s \vert e)$. Details of this computation are given in the Appendix. For example, Table \ref{table:cond-prob} shows the most likely entities in {\em park} scenes. 
For all images with at least three detected entities, we consider the image-sentence alignment probability assigned by CLIP to the {\em scene}-level description, when the top 1, 2 or 3 most likely entities in the scene are masked. Figure \ref{fig:scene-prob} displays a linear trend, with the probability assigned by CLIP to the scene-level description dropping as more likely entities are removed.\footnote{A one-way ANOVA comparing the change in log probability as 1, 2 or 3 entities are removed showed that the difference is significant ($F(2,156)=4.25, p < 0.05)$.} Thus, when CLIP aligns images with scenes, it is relying on object-level information in the visual modality. This explains why the removal of object mentions in text results in higher preference for scene-level descriptions, since the objects are detectable in the image. By the same token, masking objects in images causes the model to rely more on the entity-level information in the text.

\paragraph{Single word comparison }
So far, our analysis suggests that CLIP 
%relies on object-level information. 
reasons about scenes based on object-level information.
However, the length of the caption might be a possible confounding factor. Some of our results might simply be due to the model assigning a higher alignment probability to a caption which is longer or more informative. This would be an alternative explanation for the changes observed above in the alignment probabilities after textual ablation.
%model might prefer the caption with more information rather than the more grounded. 

To account for this, we replicate the alignment experiment using single words. We use the scene labels extracted from HL1K scene-level descriptions and identify the top three most likely entities in a given scene, as identified in the previous experiment (see Figure \ref{fig:scene-prob}). Given an image, we compare image-text alignment probabilities in CLIP for single-word object labels (e.g. \textit{motorbike}) and single-word scene labels (e.g. \textit{road}). 
%For object-level captions, we use a noun corresponding to one of the entities mentioned (e.g. \textit{motorbike}); for scene-level captions, we simply use the scene label (\textit{road}).
% replace the objective caption with the entities mentioned, obtaining one-word scene-entity pairs (e.g \textit{'road', 'motorbike'}).

\begin{figure}[!t]
    \begin{minipage}[b]{\linewidth}
        \centering
        \includegraphics[scale=0.3]{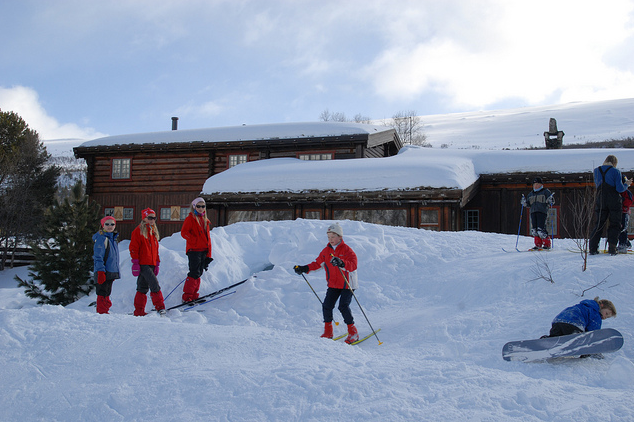}
        \caption*{\textit{resort}: 3\% \textit{person}: 96\% }
    \end{minipage}
    \begin{minipage}[b]{\linewidth}
        \centering
        \includegraphics[scale=0.3]{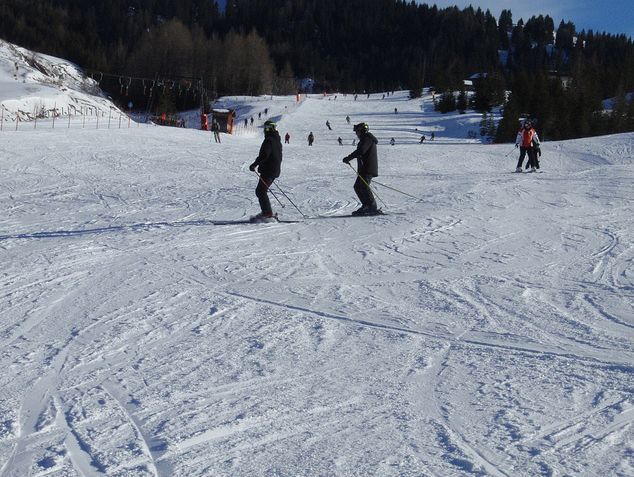}
        \caption*{\textit{resort}: 99\% \textit{snowboard}: 1\% }
    \end{minipage}
    \caption{Scene vs entity one-to-one comparison. In the top image, there are many people in the foreground and the entity \textit{person} is preferred over the scene label \textit{resort}. At the bottom, people are snowboarding in the background and the scene label is preferred over the the entity (\textit{snowboard}).  }
    \label{fig:one-word-example}
\end{figure}

In this setting, CLIP displays a moderate preference for scene labels (63\%), suggesting that such labels are more informative than object-level labels, for the one-word alignment task. Moreover, a qualitative analysis shows that CLIP prefers object labels when the images have strongly foregrounded entities with little background visibility, as shown in Fig~\ref{fig:one-word-example}.

\section{Related work}

Large, pretrained models are often analysed via probe tasks or through an investigation of their attention heads \cite[see][for a survey]{Belinkov2019}. For example, \citet{Li2020} consider VisualBERT's attention heads in a manner similar to \citet{Clark2019}, showing that it is able to ground entities and syntactic relations \cite[see also][]{Ilharco2020,dahlgren-lindstrom-etal-2020-probing}. \citet{Hendricks2021} similarly seek to obtain an in-depth understanding of the representations learned by V\&L models, finding that they fail to ground verbs in visual data, compared to other morphosyntactic categories.
% KvD Might it be useful to spell out briefly how this is similar to what we've done? ("Our work can be seen as applying a kind of probing to ..."). At the moment, the relevance of Belinkov, Clark, etc. is left a bit implicit. (Shows my ignorance!)

The ability of V\&L models to reason with a combination of linguistic and visual cues is explored through tasks such as VCR \cite{zellers2019recognition}, SWAG \cite{Zellers2018} and NLVR \cite{Suhr2017,Suhr2019}. \citet{Pezzelle2020}, in work complementary to our own, address the relationship between visual and textual modalities, exploring a task in which the text does not provide an object-level description of an image.  In work presented concurrently with our own, \citet{Frank2021} propose an ablation-based methodology to evaluate the extent to which multimodal models rely on visual and/or textual information to perform grounding. 

Although scene recognition is a central task in computer vision, there has been little work exploring the capabilities of models to link scene- and object-level representations. Some precedent for the concerns addressed in this paper are found in the image captioning literature. For example, an influential proposal by \cite{anderson2018bottom} combines top-down and bottom-up attention to improve the quality of image descriptions, while CapWAP \cite{fisch-etal-2020-capwap} conditions image captioning on questions that determine which information is relevant to current communicative needs, going beyond object-level description. Closer to the scope of the work presented here, a recent pretrained V\&L model, SemVLP \cite{Li2021}, combines single- and dual-streams for feature-level and high-level semantic alignment. We plan to investigate this model further in future work.

\section{Conclusions}\label{sec:conclusion}

% KvD This Conclusion focuses entirely on an existing model, CLIP. It's true that the huge difference in performance between CLIP and the other models needs discussing (and it seems to me we don't fully understand it, do we?)

% KvD But should the Conclusion not also (and perhaps even mainly) highlight what we've learnt from our ablation experiments?

% This paper considered the ability of pretrained V\&L models to align object-level and scene-level descriptions with images. A description conveying the type of scene gives less detailed information than one mentioning objects and their relations; nevertheless, scene knowledge generates expectations about objects and their configuration. 
% To ground 
Grounding language in vision requires
%and reason with both modalities, 
V\&L models to capture the relation between `high-level' characterisations of scenes and the entities they contain. The experiments in this paper suggest that when models do this, they rely on object-level information in the {\em visual} modality, to link images to scene descriptions in the {\em textual} modality. This is partly dependent on the probability of entities occurring in scenes.

However, this is not an ability that all models have: LXMERT and VisualBERT
perform poorly on scene-level descriptions and also on object-level captions which are stylistically different from the ones they were pretrained on, whereas CLIP handles both object- and scene-level captions. We note that for LXMERT and VisualBERT, testing on ADE20k, Localized Narratives and HL1K amounts to a zero-shot setting. With the exception of HL1K, a new dataset, it is an open question whether this is true of CLIP. Since this model is trained on web-scale data, it is difficult to ascertain that the training and test data in our experiments are disjoint; as \citet{Bender2021} recently argued, with such pre-training strategies, training data is often unfathomable. On the other hand, our findings also show that, at least as far as object and scene-level grounding is concerned, model size is not a determining factor. As shown in Table \ref{table:models-comparison}, CLIP is smaller in terms of number of parameters than LXMERT.

Apart from the sheer volume of pre-training data used in CLIP, we believe that two additional factors contribute to its success. First, its contrastive learning objective may result in greater sensitivity to fine-grained distinctions between captions, when the model computes the likelihood with which a text matches a given image. A second important feature of CLIP is its visual backbone, which (in the version used in this paper) is based on Visual Transformer \cite[ViT][]{Dosovitskiy2020}. By contrast, the visual component in LXMERT and VisualBERT is CNN-based. Recent work by \citet{Tuli2021} has shown that the attention-based ViT architecture may be more consistent with characteristics of human vision than a convolutional network. This could partially underlie the model's ability to use object-level information in an image to align to scene-level captions. The extent to which the visual backbone of V\&L models impacts their grounding capabilities is a topic that should be further explored.

% At the same time, we note that for LXMERT and VisualBERT, testing on ADE20k, Localized Narratives and HL1K amount to a zero-shot setting. With the exception of HL1K, a new dataset, it is an open question whether this is true of CLIP: as \citet{Bender2021} recently argued, web-scale training data tends to be unfathomable. 

\section{Ethical considerations}
For the studies presented here, we used a new dataset, HL1K, collected using the Amazon Mechanical Turk crowdsourcing platform. For the data collection, participants were shown images and asked to answer questions such as {\em Where is the picture taken?} Answers took the form of short statements. Workers were paid at the rate of \euro{0.03} per item, an amount we consider equitable for the work involved, and in line with rates for similar tasks. No sensitive or identifying information was collected.
All other data and models used are publicly available. The HL1K dataset will be made available upon publication.

%Uncomment for camera-ready
\section*{Acknowledgments}
 This project has received funding from the European Union’s Horizon 2020 research and innovation programme under the Marie
Skłodowska-Curie grant agreement No 860621.
 
\bibliography{anthology, refs}

\begin{thebibliography}{52}
\expandafter\ifx\csname natexlab\endcsname\relax\def\natexlab#1{#1}\fi

\bibitem[{Anderson et~al.(2018)Anderson, He, Buehler, Teney, Johnson, Gould,
  and Zhang}]{anderson2018bottom}
Peter Anderson, Xiaodong He, Chris Buehler, Damien Teney, Mark Johnson, Stephen
  Gould, and Lei Zhang. 2018.
\newblock Bottom-up and top-down attention for image captioning and visual
  question answering.
\newblock In \emph{Proceedings of the IEEE conference on computer vision and
  pattern recognition}, pages 6077--6086.

\bibitem[{Belinkov and Glass(2019)}]{Belinkov2019}
Yonatan Belinkov and James Glass. 2019.
\newblock \href {https://doi.org/10.1162/tacl_a_00254} {{Analysis Methods in
  Neural Language Processing: A Survey}}.
\newblock \emph{Transactions of the Association for Computational Linguistics},
  7:49--72.

\bibitem[{Bender et~al.(2021)Bender, Gebru, McMillan-Major, and
  Shmitchell}]{Bender2021}
Emily~M Bender, Timnit Gebru, Angelina McMillan-Major, and Shmargaret
  Shmitchell. 2021.
\newblock {On the Dangers of Stochastic Parrots: Can Language Models Be Too
  Big?}
\newblock In \emph{Proceedings of the fourth ACM Conference on Fairness,
  Accountability, and Transparency (FAccT'21)}, Online. Association for
  Computing Machinery.

\bibitem[{Bender and Koller(2020)}]{bender-koller-2020-climbing}
Emily~M. Bender and Alexander Koller. 2020.
\newblock \href {https://doi.org/10.18653/v1/2020.acl-main.463} {Climbing
  towards {NLU}: {On} meaning, form, and understanding in the age of data}.
\newblock In \emph{Proceedings of the 58th Annual Meeting of the Association
  for Computational Linguistics}, pages 5185--5198, Online. Association for
  Computational Linguistics.

\bibitem[{Biederman et~al.(1982)Biederman, Mezzanotte, and
  Rabinowitz}]{Biederman1982}
Irving Biederman, Robert~J. Mezzanotte, and Jan~C. Rabinowitz. 1982.
\newblock \href {https://doi.org/10.1016/0010-0285(82)90007-X} {{Scene
  perception: Detecting and judging objects undergoing relational violations}}.
\newblock \emph{Cognitive Psychology}, 14(2):143--177.

\bibitem[{Bisk et~al.(2020)Bisk, Holtzman, Thomason, Andreas, Bengio, Chai,
  Lapata, Lazaridou, May, Nisnevich, Pinto, and
  Turian}]{bisk-etal-2020-experience}
Yonatan Bisk, Ari Holtzman, Jesse Thomason, Jacob Andreas, Yoshua Bengio, Joyce
  Chai, Mirella Lapata, Angeliki Lazaridou, Jonathan May, Aleksandr Nisnevich,
  Nicolas Pinto, and Joseph Turian. 2020.
\newblock \href {https://doi.org/10.18653/v1/2020.emnlp-main.703} {Experience
  grounds language}.
\newblock In \emph{Proceedings of the 2020 Conference on Empirical Methods in
  Natural Language Processing (EMNLP)}, pages 8718--8735, Online. Association
  for Computational Linguistics.

\bibitem[{Brown et~al.(2020)Brown, Mann, Ryder, Subbiah, Kaplan, Dhariwal,
  Neelakantan, Shyam, Sastry, Askell, Agarwal, Herbert-Voss, Krueger, Henighan,
  Child, Ramesh, Ziegler, Wu, Winter, Hesse, Chen, Sigler, Litwin, Gray, Chess,
  Clark, Berner, McCandlish, Radford, Sutskever, and Amodei}]{Brown2020}
Tom~B. Brown, Benjamin Mann, Nick Ryder, Melanie Subbiah, Jared Kaplan,
  Prafulla Dhariwal, Arvind Neelakantan, Pranav Shyam, Girish Sastry, Amanda
  Askell, Sandhini Agarwal, Ariel Herbert-Voss, Gretchen Krueger, Tom Henighan,
  Rewon Child, Aditya Ramesh, Daniel~M. Ziegler, Jeffrey Wu, Clemens Winter,
  Christopher Hesse, Mark Chen, Eric Sigler, Mateusz Litwin, Scott Gray,
  Benjamin Chess, Jack Clark, Christopher Berner, Sam McCandlish, Alec Radford,
  Ilya Sutskever, and Dario Amodei. 2020.
\newblock \href {http://arxiv.org/abs/2005.14165} {{Language Models are
  Few-Shot Learners}}.
\newblock \emph{ArXiv}, 2005.14165.

\bibitem[{Bugliarello et~al.(2020)Bugliarello, Cotterell, Okazaki, and
  Elliott}]{bugliarello2020multimodal}
Emanuele Bugliarello, Ryan Cotterell, Naoaki Okazaki, and Desmond Elliott.
  2020.
\newblock Multimodal pretraining unmasked: Unifying the vision and language
  berts.
\newblock \emph{arXiv preprint arXiv:2011.15124}.

\bibitem[{Chen et~al.(2015)Chen, Fang, Lin, Vedantam, Zitnick, Gupta, and
  Doll}]{Chen2015}
Xinlei Chen, Hao Fang, Tsung-yi Lin, Ramakrishna Vedantam, C~Lawrence Zitnick,
  Saurabh Gupta, and Piotr Doll. 2015.
\newblock \href {http://arxiv.org/abs/arXiv:1504.00325v2} {{Microsoft COCO
  Captions : Data Collection and Evaluation Server}}.
\newblock \emph{arXiv}, 1504.00325:1--7.

\bibitem[{Chen et~al.(2020)Chen, Li, Yu, Kholy, Ahmed, Gan, Cheng, and
  Liu}]{chen2020uniter}
Yen-Chun Chen, Linjie Li, Licheng Yu, Ahmed~El Kholy, Faisal Ahmed, Zhe Gan,
  Yu~Cheng, and Jingjing Liu. 2020.
\newblock Uniter: Universal image-text representation learning.
\newblock In \emph{ECCV}.

\bibitem[{Clark et~al.(2019)Clark, Khandelwal, Levy, and Manning}]{Clark2019}
Kevin Clark, Urvashi Khandelwal, Omer Levy, and Christopher~D. Manning. 2019.
\newblock \href {https://doi.org/10.18653/v1/w19-4828} {{What does BERT look
  at? An analysis of BERT's attention}}.
\newblock In \emph{Proceedings of the 2019 ACL Workshop BlackboxNLP: Analyzing
  and Interpreting Neural Networks for NLP}, pages 276--286, Florence, Italy.
  Association for Computational Linguistics.

\bibitem[{Dahlgren~Lindstr{\"o}m et~al.(2020)Dahlgren~Lindstr{\"o}m,
  Bj{\"o}rklund, Bensch, and Drewes}]{dahlgren-lindstrom-etal-2020-probing}
Adam Dahlgren~Lindstr{\"o}m, Johanna Bj{\"o}rklund, Suna Bensch, and Frank
  Drewes. 2020.
\newblock \href {https://doi.org/10.18653/v1/2020.coling-main.64} {Probing
  multimodal embeddings for linguistic properties: the visual-semantic case}.
\newblock In \emph{Proceedings of the 28th International Conference on
  Computational Linguistics}, pages 730--744, Barcelona, Spain (Online).
  International Committee on Computational Linguistics.

\bibitem[{Devlin et~al.(2019)Devlin, Chang, Lee, and
  Toutanova}]{devlin2019bert}
Jacob Devlin, Ming-Wei Chang, Kenton Lee, and Kristina Toutanova. 2019.
\newblock \href {https://doi.org/10.18653/v1/N19-1423} {{BERT}: Pre-training of
  deep bidirectional transformers for language understanding}.
\newblock In \emph{Proceedings of the 2019 Conference of the North {A}merican
  Chapter of the Association for Computational Linguistics: Human Language
  Technologies, Volume 1 (Long and Short Papers)}, pages 4171--4186,
  Minneapolis, Minnesota. Association for Computational Linguistics.

\bibitem[{Dosovitskiy et~al.(2020)Dosovitskiy, Beye, Kolesnikov, Weissenborn,
  Zhai, Unterthiner, Dehghani, Minderer, Heigold, Gelly, Uszkoreit, and
  Houlsby}]{Dosovitskiy2020}
Alexey Dosovitskiy, Lucas Beye, Alexander Kolesnikov, Dirk Weissenborn, Xiaohua
  Zhai, Thomas Unterthiner, Mostafa Dehghani, Matthias Minderer, Georg Heigold,
  Sylvain Gelly, Jakob Uszkoreit, and Neil Houlsby. 2020.
\newblock {An image is worth 16x16 words: Transformers for image recognitoin at
  scale}.
\newblock \emph{arXiv}, 2010.11929.

\bibitem[{Fisch et~al.(2020)Fisch, Lee, Chang, Clark, and
  Barzilay}]{fisch-etal-2020-capwap}
Adam Fisch, Kenton Lee, Ming-Wei Chang, Jonathan Clark, and Regina Barzilay.
  2020.
\newblock \href {https://doi.org/10.18653/v1/2020.emnlp-main.705} {{C}ap{WAP}:
  Image captioning with a purpose}.
\newblock In \emph{Proceedings of the 2020 Conference on Empirical Methods in
  Natural Language Processing (EMNLP)}, pages 8755--8768, Online. Association
  for Computational Linguistics.

\bibitem[{Frank et~al.(2021)Frank, Bugliarello, and Elliott}]{Frank2021}
Stella Frank, Emanuele Bugliarello, and Desmond Elliott. 2021.
\newblock \href {http://arxiv.org/abs/2109.04448} {{Vision-and-Language or
  Vision-for-Language? On Cross-Modal Influence in Multimodal Transformers}}.
\newblock \emph{ArXiv}, 2109.04448.

\bibitem[{Harnad(1990)}]{Harnad1990}
Stevan Harnad. 1990.
\newblock \href
  {http://www.sciencedirect.com/science/article/pii/0167278990900876} {{The
  symbol grounding problem}}.
\newblock \emph{Physica}, D42(1990):335--346.

\bibitem[{Hendricks and Nematzadeh(2021)}]{Hendricks2021}
Lisa~Anne Hendricks and Aida Nematzadeh. 2021.
\newblock \href {http://arxiv.org/abs/2106.09141} {{Probing Image-Language
  Transformers for Verb Understanding}}.
\newblock \emph{ArXiv}, 2106.09141.

\bibitem[{Huang et~al.(2021)Huang, Zeng, Huang, Liu, Fu, and Fu}]{Huang2021}
Zhicheng Huang, Zhaoyang Zeng, Yupan Huang, Bei Liu, Dongmei Fu, and Jianlong
  Fu. 2021.
\newblock \href {http://arxiv.org/abs/2104.03135} {{Seeing Out of tHe bOx:
  End-to-End Pre-training for Vision-Language Representation Learning}}.
\newblock \emph{ArXiv}, 2104.03135.

\bibitem[{Ilharco et~al.(2020)Ilharco, Zellers, Farhadi, and
  Hajishirzi}]{Ilharco2020}
Gabriel Ilharco, Rowan Zellers, Ali Farhadi, and Hannaneh Hajishirzi. 2020.
\newblock \href {http://arxiv.org/abs/arXiv:2005.00619v4} {{Probing Contextual
  Language Models for Common Ground with Visual Representations}}.
\newblock \emph{arXiv}, 2005.00619.

\bibitem[{Itti and Koch(2001)}]{Itti2001}
L~Itti and C~Koch. 2001.
\newblock \href {https://doi.org/10.1038/35058500} {{Computational modelling of
  visual attention.}}
\newblock \emph{Nature reviews neuroscience}, 2(3):194--203.

\bibitem[{Jayanthi et~al.(2020)Jayanthi, Pruthi, and
  Neubig}]{jayanthi2020neuspell}
Sai~Muralidhar Jayanthi, Danish Pruthi, and Graham Neubig. 2020.
\newblock Neuspell: A neural spelling correction toolkit.
\newblock \emph{arXiv preprint arXiv:2010.11085}.

\bibitem[{Kim et~al.(2021)Kim, Son, and Kim}]{kim2021vilt}
Wonjae Kim, Bokyung Son, and Ildoo Kim. 2021.
\newblock Vilt: Vision-and-language transformer without convolution or region
  supervision.
\newblock \emph{arXiv preprint arXiv:2102.03334}.

\bibitem[{Li et~al.(2021)Li, Yan, Xu, Luo, Wang, Bi, and Huang}]{Li2021}
Chenliang Li, Ming Yan, Haiyang Xu, Fuli Luo, Wei Wang, Bin Bi, and Songfang
  Huang. 2021.
\newblock {SemVLP: Vision-Language Pre-training by Aligning Semantics at
  Multiple Levels}.
\newblock \emph{arXiv preprint}, 2103.07829.

\bibitem[{Li et~al.(2020{\natexlab{a}})Li, Duan, Fang, Gong, and
  Jiang}]{Li-etal-2020unicodervl}
Gen Li, Nan Duan, Yuejian Fang, Ming Gong, and Daxin Jiang. 2020{\natexlab{a}}.
\newblock \href {https://aaai.org/ojs/index.php/AAAI/article/view/6795}
  {Unicoder-vl: {A} universal encoder for vision and language by cross-modal
  pre-training}.
\newblock In \emph{The Thirty-Fourth {AAAI} Conference on Artificial
  Intelligence, {AAAI} 2020, The Thirty-Second Innovative Applications of
  Artificial Intelligence Conference, {IAAI} 2020, The Tenth {AAAI} Symposium
  on Educational Advances in Artificial Intelligence, {EAAI} 2020, New York,
  NY, USA, February 7-12, 2020}, pages 11336--11344. {AAAI} Press.

\bibitem[{Li et~al.(2019)Li, Yatskar, Yin, Hsieh, and Chang}]{li2019visualbert}
Liunian~Harold Li, Mark Yatskar, Da~Yin, Cho-Jui Hsieh, and Kai-Wei Chang.
  2019.
\newblock Visualbert: A simple and performant baseline for vision and language.
\newblock In \emph{Arxiv}.

\bibitem[{Li et~al.(2020{\natexlab{b}})Li, Yatskar, Yin, Hsieh, and
  Chang}]{Li2020}
Liunian~Harold Li, Mark Yatskar, Da~Yin, Cho-Jui Hsieh, and Kai-Wei Chang.
  2020{\natexlab{b}}.
\newblock \href {https://doi.org/10.18653/v1/2020.acl-main.469} {{What Does
  BERT with Vision Look At?}}
\newblock In \emph{Proceedings ofthe 58th Annual Meeting ofthe Association for
  Computational Linguistics (ACL'20)}, pages 5265--5275, Online. Association
  for Computational Linguistics.

\bibitem[{Li et~al.(2020{\natexlab{c}})Li, Yin, Li, Zhang, Hu, Zhang, Wang, Hu,
  Dong, Wei et~al.}]{li2020oscar}
Xiujun Li, Xi~Yin, Chunyuan Li, Pengchuan Zhang, Xiaowei Hu, Lei Zhang, Lijuan
  Wang, Houdong Hu, Li~Dong, Furu Wei, et~al. 2020{\natexlab{c}}.
\newblock Oscar: Object-semantics aligned pre-training for vision-language
  tasks.
\newblock In \emph{European Conference on Computer Vision}, pages 121--137.
  Springer.

\bibitem[{Lin et~al.(2014)Lin, Maire, Belongie, Hays, Perona, Ramanan,
  Doll{\'a}r, and Zitnick}]{lin2014microsoft}
Tsung-Yi Lin, Michael Maire, Serge Belongie, James Hays, Pietro Perona, Deva
  Ramanan, Piotr Doll{\'a}r, and C~Lawrence Zitnick. 2014.
\newblock Microsoft coco: Common objects in context.
\newblock In \emph{European conference on computer vision}, pages 740--755.
  Springer.

\bibitem[{Lu et~al.(2019)Lu, Batra, Parikh, and Lee}]{lu2019vilbert}
Jiasen Lu, Dhruv Batra, Devi Parikh, and Stefan Lee. 2019.
\newblock Vilbert: Pretraining task-agnostic visiolinguistic representations
  for vision-and-language tasks.
\newblock In \emph{Advances in Neural Information Processing Systems}, pages
  13--23.

\bibitem[{Lu et~al.(2020)Lu, Goswami, Rohrbach, Parikh, and
  Lee}]{lu2020vilbert12in1}
Jiasen Lu, Vedanuj Goswami, Marcus Rohrbach, Devi Parikh, and Stefan Lee. 2020.
\newblock 12-in-1: Multi-task vision and language representation learning.
\newblock In \emph{The IEEE/CVF Conference on Computer Vision and Pattern
  Recognition (CVPR)}.

\bibitem[{Luo et~al.(2020)Luo, Ji, Shi, Huang, Duan, Li, Chen, and
  Zhou}]{Luo2020}
Huaishao Luo, Lei Ji, Botian Shi, Haoyang Huang, Nan Duan, Tianrui Li, Xilin
  Chen, and Ming Zhou. 2020.
\newblock \href {http://arxiv.org/abs/2002.06353} {{UniVL: A unified video and
  language pre-training model for multimodal understanding and generation}}.
\newblock \emph{arXiv}.

\bibitem[{Ma et~al.(2010)Ma, Zhu, Lyu, and King}]{Ma2010}
Hao Ma, Jianke Zhu, Michael Rung~Tsong Lyu, and Irwin King. 2010.
\newblock \href {https://doi.org/10.1109/TMM.2010.2051360} {{Bridging the
  semantic gap between image contents and tags}}.
\newblock \emph{IEEE Transactions on Multimedia}, 12(5):462--473.

\bibitem[{Malcolm et~al.(2016)Malcolm, Groen, and Baker}]{Malcolm2016}
George~L. Malcolm, Iris~I.A. Groen, and Chris~I. Baker. 2016.
\newblock \href {https://doi.org/10.1016/j.tics.2016.09.003} {{Making Sense of
  Real-World Scenes}}.
\newblock \emph{Trends in Cognitive Sciences}, 20(11):843--856.

\bibitem[{Oliva and Torralba(2007)}]{Oliva2007}
Aude Oliva and Antonio Torralba. 2007.
\newblock \href {https://doi.org/10.1016/j.tics.2007.09.009} {{The role of
  context in object recognition.}}
\newblock \emph{Trends in cognitive sciences}, 11(12):520--527.

\bibitem[{Parcabalescu et~al.(2021)Parcabalescu, Gatt, Frank, and
  Calixto}]{Parcabalescu2021}
Letitia Parcabalescu, Albert Gatt, Annette Frank, and Iacer Calixto. 2021.
\newblock \href {https://arxiv.org/abs/2012.12352} {Seeing past words: Testing
  the cross-modal capabilities of pretrained v\&l models on counting tasks}.
\newblock In \emph{Proceedings of the Workshop Beyond Language: Multimodal
  Semantic Representations (MMSR'21)}, Groningen, The Netherlands.

\bibitem[{Pezzelle et~al.(2020)Pezzelle, Greco, Gandolfi, Gualdoni, and
  Bernardi}]{Pezzelle2020}
Sandro Pezzelle, Claudio Greco, Greta Gandolfi, Eleonora Gualdoni, and
  Raffaella Bernardi. 2020.
\newblock \href {https://doi.org/10.18653/v1/2020.findings-emnlp.248} {{Be
  Different to Be Better! A Benchmark to Leverage the Complementarity of
  Language and Vision}}.
\newblock In \emph{Findings ofthe Association for Computational Linguistics:
  EMNLP 2020}, pages 2751--2767, Online. Association for Computational
  Linguistics.

\bibitem[{Pont-Tuset et~al.(2019)Pont-Tuset, Uijlings, Changpinyo, Soricut, and
  Ferrari}]{Pont-Tuset2019}
Jordi Pont-Tuset, Jasper Uijlings, Soravit Changpinyo, Radu Soricut, and
  Vittorio Ferrari. 2019.
\newblock \href {http://arxiv.org/abs/1912.03098} {{Connecting Vision and
  Language with Localized Narratives}}.
\newblock \emph{arXiv}, 1912.03098.

\bibitem[{Radford et~al.(2021)Radford, Kim, Hallacy, Ramesh, Goh, Agarwal,
  Sastry, Askell, Mishkin, Clark, Krueger, and Sutskever}]{Radford2021}
Alec Radford, Jong~Wook Kim, Chris Hallacy, Aditya Ramesh, Gabriel Goh,
  Sandhini Agarwal, Girish Sastry, Amanda Askell, Pamela Mishkin, Jack Clark,
  Gretchen Krueger, and Ilya Sutskever. 2021.
\newblock \href {http://arxiv.org/abs/2103.00020} {{Learning Transferable
  Visual Models From Natural Language Supervision}}.
\newblock \emph{arXiv preprint}, 2103.00020.

\bibitem[{Ren et~al.(2015)Ren, He, Girshick, and Sun}]{Ren2015}
Shaoqing Ren, Kaiming He, Ross Girshick, and Jian Sun. 2015.
\newblock {Faster R-CNN: Towards Real-Time Object Detection with Region
  Proposal Networks}.
\newblock In \emph{Advances in Neural Information Processing Systems 28
  (NeurIPS 2015)}, Montreal, Canada.

\bibitem[{Su et~al.(2020)Su, Zhu, Cao, Li, Lu, Wei, and Dai}]{Su2020VL-BERT}
Weijie Su, Xizhou Zhu, Yue Cao, Bin Li, Lewei Lu, Furu Wei, and Jifeng Dai.
  2020.
\newblock \href {https://openreview.net/forum?id=SygXPaEYvH} {Vl-bert:
  Pre-training of generic visual-linguistic representations}.
\newblock In \emph{International Conference on Learning Representations}.

\bibitem[{Suhr et~al.(2017)Suhr, Lewis, Yeh, and Artzi}]{Suhr2017}
Alane Suhr, Mike Lewis, James Yeh, and Yoav Artzi. 2017.
\newblock \href {https://doi.org/10.18653/v1/P17-2034} {{A Corpus of Natural
  Language for Visual Reasoning}}.
\newblock In \emph{Proceedings ofthe 55th Annual Meeting ofthe Association for
  Computational Linguistics (ACL'17)}, pages 217--223, Vancouver, BC.
  Association for Computational Linguistics.

\bibitem[{Suhr et~al.(2019)Suhr, Zhou, Zhang, Zhang, Bai, and Artzi}]{Suhr2019}
Alane Suhr, Stephanie Zhou, Ally Zhang, Iris Zhang, Huajun Bai, and Yoav Artzi.
  2019.
\newblock \href {https://doi.org/10.18653/v1/p19-1644} {{A Corpus for Reasoning
  about Natural Language Grounded in Photographs}}.
\newblock \emph{arXiv}, 1811.00491:6418--6428.

\bibitem[{Tan and Bansal(2019)}]{tan-bansal-2019-lxmert}
Hao Tan and Mohit Bansal. 2019.
\newblock \href {https://doi.org/10.18653/v1/D19-1514} {{LXMERT}: Learning
  cross-modality encoder representations from transformers}.
\newblock In \emph{Proceedings of the 2019 Conference on Empirical Methods in
  Natural Language Processing and the 9th International Joint Conference on
  Natural Language Processing (EMNLP-IJCNLP)}, pages 5100--5111, Hong Kong,
  China. Association for Computational Linguistics.

\bibitem[{Torralba et~al.(2006)Torralba, Oliva, Castelhano, and
  Henderson}]{Torralba2006}
Antonio Torralba, Aude Oliva, Monica~S. Castelhano, and John~M. Henderson.
  2006.
\newblock \href {https://doi.org/10.1037/0033-295X.113.4.766} {{Contextual
  guidance of eye movements and attention in real-world scenes: The role of
  global features in object search.}}
\newblock \emph{Psychological Review}, 113(4):766--786.

\bibitem[{Tuli et~al.(2021)Tuli, Dasgupta, Grant, and Griffiths}]{Tuli2021}
Shikhar Tuli, Ishita Dasgupta, Erin Grant, and Thomas~L. Griffiths. 2021.
\newblock \href {http://arxiv.org/abs/2105.07197} {{Are Convolutional Neural
  Networks or Transformers more like human vision?}}
\newblock \emph{arXiv}, 2105.07197.

\bibitem[{V{\~{o}} and Wolfe(2013)}]{Vo2013}
Melissa~L.H. V{\~{o}} and Jeremy~M. Wolfe. 2013.
\newblock \href {https://doi.org/10.1177/0956797613476955} {{Differential
  Electrophysiological Signatures of Semantic and Syntactic Scene Processing}}.
\newblock \emph{Psychological Science}, 24(9):1816--1823.

\bibitem[{Wang et~al.(2020)Wang, Hu, Zhang, Li, Wang, Zhang, Gao, and
  Liu}]{Wang2020}
Jianfeng Wang, Xiaowei Hu, Pengchuan Zhang, Xiujun Li, Lijuan Wang, Lei Zhang,
  Jianfeng Gao, and Zicheng Liu. 2020.
\newblock \href {http://arxiv.org/abs/2012.06946} {{MiniVLM: A Smaller and
  Faster Vision-Language Model}}.
\newblock \emph{arXiv}, 2012.06946.

\bibitem[{Zellers et~al.(2019)Zellers, Bisk, Farhadi, and
  Choi}]{zellers2019recognition}
Rowan Zellers, Yonatan Bisk, Ali Farhadi, and Yejin Choi. 2019.
\newblock From recognition to cognition: Visual commonsense reasoning.
\newblock In \emph{Proceedings of the IEEE/CVF Conference on Computer Vision
  and Pattern Recognition}, pages 6720--6731.

\bibitem[{Zellers et~al.(2018)Zellers, Bisk, Schwartz, and Choi}]{Zellers2018}
Rowan Zellers, Yonatan Bisk, Roy Schwartz, and Yejin Choi. 2018.
\newblock \href {https://doi.org/arXiv:1808.05326v1} {{SWAG: A Large-Scale
  Adversarial Dataset for Grounded Commonsense Inference}}.
\newblock \emph{arXiv}, 1808.05326.

\bibitem[{Zhou et~al.(2017)Zhou, Zhao, Puig, Fidler, Barriuso, and
  Torralba}]{zhou2017scene}
Bolei Zhou, Hang Zhao, Xavier Puig, Sanja Fidler, Adela Barriuso, and Antonio
  Torralba. 2017.
\newblock Scene parsing through ade20k dataset.
\newblock In \emph{Proceedings of the IEEE conference on computer vision and
  pattern recognition}, pages 633--641.

\bibitem[{Zhu and Bhat(2020)}]{Zhu2020}
Wanzheng Zhu and Suma Bhat. 2020.
\newblock \href {https://doi.org/10.18653/v1/2020.findings-emnlp.9} {{GRUEN for
  evaluating linguistic quality of generated text}}.
\newblock \emph{arXiv}, 2010.02498.

\end{thebibliography}
\bibliographystyle{acl_natbib}
\clearpage

\appendix
\section{Dataset details}\label{sec:datastats}
Table \ref{table:datasets-details} gives the sizes of the ADE20k,  HL1K and COCO datasets. For ADE20k, the numbers are for images which are not labelled as having an {\em unknown} scene. For COCO captions, there are five captions associated with each image. For the purposes of the present study, a single caption is randomly selected for each.

\begin{table}[h]
    \small
    \centering
    \begin{tabular}{c|c|c|c}
        \toprule
        & {ADE20k} & {HL1K}  & {COCO} \\
        \midrule
        images & 19733 & 1000 & 5000\\
        \midrule
        COCO captions & 19733 & 3000 & 5000\\
        \midrule
        Localized Narratives & 19733 & 1000 & 5000\\
        \bottomrule
    \end{tabular}
    \caption{Datasets sizes}
     \label{table:datasets-details}
\end{table}

Table \ref{table:datasets-details-ablation} provides the number of ablations analysed in the study. For both ADE20k and HL1k we obtain a number of ablated captions which is greater than the respective dataset sizes in Table \ref{table:datasets-details}, because for each example we generate all the possible combinations of noun phrases. For the Visual and Visual+Textual ablations, the number of ablated instances is lower than the dataset size, because we skip all the images where no object is detected.

\begin{table}[h]
    \small
    \centering
    \begin{tabular}{c|c|c}
        \toprule
        & {ADE20k} & {HL1K} \\
        \midrule
        T & 205k & 10027 \\
        \midrule
        V & 10788 & 625 \\
        \midrule
        T+V & 1078 & 625 \\
        \bottomrule
    \end{tabular}
    \caption{Total number of ablations generated per dataset, across all the ablations experiments}
     \label{table:datasets-details-ablation}
\end{table}

\subsection{Scene description templates}\label{sec:templates}
ADE20K scene labels are converted to textual descriptions using three different templates:

\begin{itemize}
    \item {\em it is a} {\sc scene}
    \item {\em this is a} {\sc scene}
    \item {\em it is located in} {\sc scene}
\end{itemize}

A scene label is rendered into a description using a randomly selected template from the above. For example:

\begin{itemize}
    \item {\em bathroom} --> it is a bathroom
    \item {\em bedroom} --> this is a bedroom
    \item {\em airport} --> it is located in an airport
\end{itemize}

\subsection{HL1K Examples}\label{sec:hl1k}
HL1K is collected by asking crowd workers to consider an image, and answer the question {\em Where is the picture taken?} Crowd workers are asked to respond using full sentences, as far as possible. Since images are selected from the COCO 2014 {\tt train} split, they are also accompanied by their object-level COCO captions and Localized Narratives. An example image with corresponding HL1K scene-level descriptions is shown below.

\begin{figure}[h]
    \centering
    \includegraphics[scale=0.33]{imgs/bedroom.jpg}

    \begin{minipage}[b]{\textwidth}
    {\bf Where is the picture taken?}
    \begin{itemize}
    \item in a bedroom
    \item the picture is taken in a bedroom
    \item this is the bedroom
    \end{itemize}
    \end{minipage}
    \caption{COCO image with corresponding scene descriptions in HL1K}
    \label{fig:my_label}
\end{figure}

\section{Experiment details}\label{sec:expdetails}
\subsection{Image-sentence alignment}
Results for image-sentence alignment experiments in the paper are averages over three separate runs for each model.

For LXMERT, we use the image-sentence alignment head.\footnote{\url{github.com/huggingface/transformers}} CLIP and LXMERT are tested on the standard alignment task: given an image and either the correct caption, or a random caption, the model needs to determine whether the caption correctly aligns with the image.

We use the publicly available implementation of VisualBERT.\footnote{\url{https://github.com/uclanlp/visualbert}} The image-sentence alignment setting for this model is somewhat different, since alignment is modelled as an extension of the next-sentence prediction task in unimodal BERT. VisualBERT takes an image and a correct caption, together with a second caption, which may be correct or randomly selected. The task is to predict whether the second caption correctly aligns with the image+caption pair.

For all experiments, we truncate textual captions to a maximum length of 50 tokens, following standard practice for such models, including CLIP.

\subsection{Ablation}\label{sec:ablation-examples}
\begin{figure}[!t]
    \begin{minipage}[b]{\linewidth}
        \centering
        \includegraphics[scale=0.45]{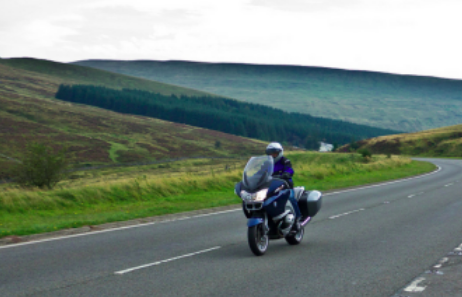}
        \caption*{Original Image}
    \end{minipage}
    \begin{minipage}[b]{\linewidth}
        \centering
        \includegraphics[scale=0.45]{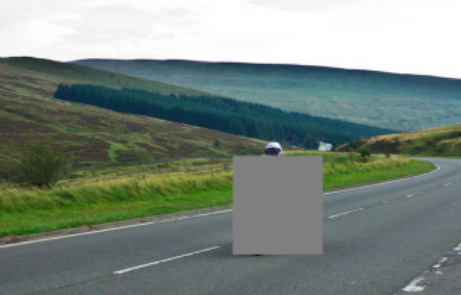}
        \caption*{Occluded Image}
    \end{minipage}
    % \vspace{5mm}
    \begin{minipage}[b]{\linewidth}
        \footnotesize
        {\bf COCO:} A man rides a motorcycle on a road through a grassy, hilly area. \\\\
        {\bf Ablated Captions:} \\
        \begin{itemize}
            \item a grassy, hilly area \textit{(A man, a motorcycle, a road)}
            \item a road \textit{(A man, a motorcycle, a grassy, hilly area)}
            \item a road through a grassy, hilly area  \textit{(A man, a motorcycle)}
            \item A man rides a road \textit{(a motorcycle, a grassy, hilly area)}
            \item a motorcycle a grassy, hilly area \textit{(A man, a road)}
            \item A man rides a grassy, hilly area \textit{(a motorcycle, a road)}
            \item A man rides a motorcycle \textit{(a road, a grassy, hilly area)}
            \item A man rides a road through a grassy, hilly area \textit{(a motorcycle)}
            \item a motorcycle on a road \textit{(A man, a grassy, hilly area)}
            \item A man rides a motorcycle on a road \textit{(a grassy, hilly area)}
            \item A man rides a motorcycle a grassy, hilly area \textit{(a road)}
            \item a motorcycle on a road through a grassy, hilly area. \textit{(A man)}
        \end{itemize}
    \end{minipage}
    \caption{The original image at the top, in the middle the visually ablated imaged and at the bottom the original caption and all the cases generated by the constituent textual ablation. } %We generate ablations for all the possible constituents combinations, although this method produces grammatically acceptable captions, they do not always make sense.}
    \label{fig:visual-textual-ablation-example}
\end{figure}

Ablation is performed on object-level captions, and on images. Results for ablation are reported based on a single run of CLIP.
Figure~\ref{fig:visual-textual-ablation-example} shows an original image and its object-level caption, together with the visual  and textual ablations. Below, we explain the process of ablation in more detail.

\paragraph{Textual ablation} Given an object-level caption, we identify all the NPs in the caption and create new versions by removing each possible subset of the set of NPs, with the restriction that the resulting caption must always contain at least one NP. When NP removal results in dangling predicates, we remove them to preserve grammaticality. NPs are detected using Spacy v.3 pipeline for English using the {\tt en\_core\_web\_md} pretrained models. 

\paragraph{Visual ablation} Given an object-level caption and an image, we extract all nouns from the caption and extract the embedding vector for each noun using the pretrained FastText embeddings.\footnote{We use the model with 2m word vectors trained with subword information from common Crawl \url{https://fasttext.cc/docs/en/english-vectors.html}} We pass the image through the Faster-RCNN object detector\footnote{Faster R-CNN ResNet-50 FPN pre-trained on COCO, available from the {\tt torchvision} module in {\tt Pytorch}} to detect entities. We extract embeddings for each entity label. Then, we identify regions to be masked by comparing embeddings for entity labels $l_e$ against embeddings for nouns $n_e$ in the caption, considering them a match if $\text{cosine}(l_e, n_e) \geq 0.7$. Bounding box regions corresponding to matched entities are occluded with a greyscale mask.

\section{Scene-entity probabilities}

\begin{figure}[!h]
    \centering
     \begin{subfigure}[b]{0.45\textwidth}
         \centering
         \includegraphics[scale=0.6]{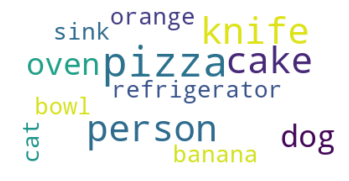}
         \caption{Scene: {\em kitchen}}
         \label{fig:kitchen}
     \end{subfigure}
     \hfill
     \begin{subfigure}[b]{0.45\textwidth}
         \centering
         \includegraphics[scale=0.6]{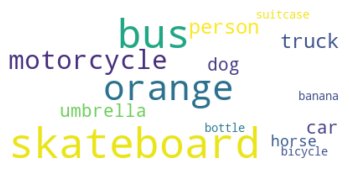}
         \caption{Scene: {\em road}}
         \label{fig:road}
     \end{subfigure}
     \hfill
     \begin{subfigure}[b]{0.45\textwidth}
         \centering
         \includegraphics[scale=0.6]{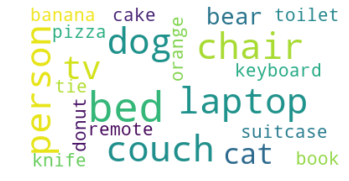}
         \caption{Scene: {\em room}}
         \label{fig:room}
     \end{subfigure}
    \hfill
    \begin{subfigure}[b]{0.45\textwidth}
         \centering
         \includegraphics[scale=0.6]{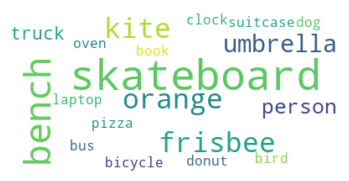}
         \caption{Scene: {\em park}}
         \label{fig:room}
     \end{subfigure}
    \caption{Visualisations of entities in four different scene types. For a given entity $e$ in scene $s$, font size is proportional to $P(s \vert e)$}
    \label{fig:wordclouds}
\end{figure}

Entity probabilities are computed on the basis of the entities detected in the image and object-level caption, using the process described above for visual ablation. Let $e$ be an entity detected $n_e$ times in the dataset, of which $n_{e,s}$ times in images depicting scene $s$. We compute:

\begin{equation*}
    P(s \vert e) = \frac{n_{e,s}}{n_e}
\end{equation*}

Figure \ref{fig:wordclouds} shows visualisations for entities detected in four different scene types, from the HL1K dataset. To compute the mean probabilities assigned by CLIP to scene-level descriptions after entities are visually ablated, we average over those images containing at least three detected entities (53/174 total scenes).

% \end{document}

\end{document}